\documentclass[11pt,a4paper]{article}
\pdfoutput=1
\usepackage[nohyperref]{acl2017}
\usepackage{times}
\usepackage{latexsym}

\usepackage{vntex}
\usepackage[english]{babel}
\usepackage{graphicx}
\usepackage{tabularx}
\usepackage{multirow}

\usepackage[hidelinks]{hyperref}

\usepackage{url}

\aclfinalcopy % Uncomment this line for the final submission
\def\aclpaperid{26} %  Enter the acl Paper ID here

\title{Fro{m}\ {W}ord Segmentation to POS Tagging for Vietnamese}

\author{Dat Quoc Nguyen$^1$, Thanh Vu$^2$, Dai Quoc Nguyen$^3$, Mark Dras$^1$ \and Mark Johnson$^1$\\
$^1$Department of Computing, Macquarie University, Australia \\
{\tt \{dat.nguyen, mark.dras, mark.johnson\}@mq.edu.au}\\
         $^2$NIHR Innovation Observatory, Newcastle University, United Kingdom\\
{\tt thanh.vu@newcastle.ac.uk}\\
$^3$PRaDA  Centre, Deakin University, Australia \\
         {\tt dai.nguyen@deakin.edu.au}}

\begin{document}
\maketitle
\begin{abstract}
This paper presents an empirical comparison of two strategies for Vietnamese Part-of-Speech (POS) tagging from unsegmented text: (i) a pipeline strategy where we consider the output of a word segmenter  as the input of a POS tagger, and (ii) a joint strategy where we predict a combined segmentation and POS tag for  each syllable. We also make a comparison between  state-of-the-art (SOTA) feature-based and neural network-based models. On the benchmark Vietnamese treebank \cite{nguyen-EtAl:2009:LAW-III}, experimental results  show that the pipeline strategy produces better scores of POS tagging from unsegmented text than the joint strategy, and the highest accuracy is obtained by using a  feature-based model.

%\textbf{Keywords}: Vietnamese, Word segmentation, POS tagging
\end{abstract}

\section{Introduction}

POS tagging is one of the most fundamental natural language processing (NLP) tasks. In English  where  white space is a strong indicator of word boundaries, POS tagging is an important first step towards many other NLP tasks. However, white space when written  in Vietnamese is also  used to separate syllables that constitute  words. So  for Vietnamese NLP, word segmentation is referred to as the key  first step \cite{Dien2001}.

When applying POS tagging to real-world Vietnamese text where gold word-segmentation is not available, the pipeline strategy is to first segment the text by using a word segmenter, and then feed the word-segmented text---which is the output of the word segmenter---as the input to a POS tagger. 
For example, given a written text ``thuế thu nhập  cá nhân'' (individual\textsubscript{cá\_nhân} income\textsubscript{thu\_nhập} tax\textsubscript{thuế}) consisting of 5 syllables, the word segmenter  returns 
a two-word phrase ``thuế\_thu\_nhập  cá\_nhân.''\footnote{In the traditional  underscore-based representation in Vietnamese word segmentation \protect\cite{nguyen-EtAl:2009:LAW-III}, white space is only used to separate words while underscore is used to separate syllables inside a word.} Then given the input segmented text ``thuế\_thu\_nhập  cá\_nhân'', the POS tagger returns  ``thuế\_thu\_nhập/N  cá\_nhân/N.''

A class of approaches to POS tagging from unsegmented  text that has been actively explored in other languages, such as in Chinese and Japanese, is  joint word segmentation and POS tagging \cite{zhang-clark:2008:ACLMain}. A possible joint strategy is to assign a combined segmentation and POS tag to each syllable \cite{kruengkrai-EtAl:2009:ACLIJCNLP}. For example, given the input  text  ``thuế thu nhập  cá nhân'', the joint strategy would produce ``thuế/B-N thu/I-N nhập/I-N  cá/B-N nhân/I-N'', where B refers to the beginning of a word and I refers to the  inside of a word. 
\newcite{ShaoHTN17} showed that this joint strategy gives SOTA results for   Chinese POS tagging by utilizing a BiLSTM-CNN-CRF model \cite{ma-hovy:2016:P16-1}. 

In this paper, we present the first empirical study comparing the joint and pipeline strategies for  Vietnamese POS tagging from  unsegmented text. In addition, we make a comparison  between SOTA feature-based and neural network-based models, which, to the best of our knowledge, has not done in any prior work on Vietnamese.  On the benchmark Vietnamese treebank \cite{nguyen-EtAl:2009:LAW-III}, we show that the pipeline strategy produces  better scores  than the joint strategy. We also show that the highest tagging accuracy is obtained by using a traditional feature-based model  rather than neural network-based models.

\section{Related work}

\subsection{Word segmentation}

 \newcite{Y06-1028},   
\newcite{Dinh2006} and \newcite{Tran2010} considered the Vietnamese word segmentation task as a sequence labeling task, using either a  CRF, SVM or MaxEnt model to assign each syllable a segmentation tag such as B or I.  In addition,  \newcite{Le2008}, \newcite{Pham2009} and \newcite{Tran2012}  used  the maximum matching method \cite{NanYuan91} to generate all possible segmentations for each input sentence; then to select the best segmentation, \newcite{Le2008} and \newcite{Tran2012} applied  n-gram language model while \newcite{Pham2009} employed POS information from an external POS tagger.  Later, \newcite{Liu2014} and \newcite{NguyenL2016} proposed  approaches  based on pointwise prediction,  where a binary classifier is trained to identify whether or not there is a word boundary at each point between two syllables. Furthermore, \newcite{NguyenNVDJ2017} proposed a rule-based approach which gets  the
highest results to date in terms of both segmentation accuracy
and  speed. 

\subsection{POS tagging}

Regarding Vietnamese POS tagging, \newcite{Dien-Kiem:2003:PARALLEL} projected  POS annotations from English  to Vietnamese via a bilingual corpus of word alignments. As a standard sequence labeling task, previous  research has  applied the CRF, SVM or MaxEnt model  to assign each word a POS tag \cite{4586344,Tran09,lehong00526139,Nguyen2010,Tran2010,Bach2013}. In addition, 
\newcite{NguyenNPP2011} proposed a rule-based approach to automatically construct transformation rules for POS tagging
in the form of a Ripple Down
Rules  tree \cite{ComptonJ90}, leading to a development of the RDRPOSTagger  \cite{NguyenEACL2014NPP}  which was the best system for 
the POS tagging shared task at the 2013 Vietnamese Language and Speech Processing (VLSP) workshop.

\newcite{NguyenNPP_AICom2015} and 
\newcite{JCSCE} later showed that  SOTA accuracies at {94+}\%  in the Vietnamese POS tagging task  are obtained by simply retraining  existing  English POS taggers on Vietnamese data, showing  that  the MarMoT tagger \cite{mueller-schmid-schutze:2013:EMNLP} and the Stanford POS tagger \cite{Toutanova:2003} obtain higher accuracies than   RDRPOSTagger.  %\footnote{\protect\newcite{JCSCE} used the language independent  version of RDRPOSTagger in their experiments.  RDRPOSTagger also provides a specific version for Vietnamese, which  produces 0.5\% higher accuracy than the language independent version when applied to Vietnamese.} 
\newcite{NguyenNPP_AICom2015} also showed that a simple lexicon-based approach assigning each word by its most probable POS tag gains a promising accuracy at  {91}\%. Note that both \newcite{NguyenNPP_AICom2015} and 
\newcite{JCSCE}   did not experiment with neural network models.
\newcite{PhamPNP2017b} recently applied  the BiLSTM-CNN-CRF \cite{ma-hovy:2016:P16-1}  for Vietnamese POS tagging, however, they did not experiment with SOTA feature-based models.

Previously, only \newcite{Takahashi2016} carried out joint  word segmentation and POS tagging for Vietnamese, to predicting a combined segmentation and POS tag to each syllable. In particular, \newcite{Takahashi2016}  experimented with traditional SVM- and CRF-based toolkits on a dataset of about 7k sentences and  reported results of joint prediction only, i.e., they did not compare to the pipeline strategy.   The  CoNLL 2017 shared task on Universal Dependencies (UD) parsing from raw text \cite{zeman-EtAl:2017:K17-3} provided some  results to the pipeline strategy from word segmentation to POS tagging, however, the Vietnamese dataset in the UD project is very small, consisting of 1,400 training sentences. Furthermore, \newcite{nguyen-dras-johnson:2017:K17-3} provided a pre-trained jPTDP model for joint POS tagging and dependency parsing for  Vietnamese,\footnote{\url{https://drive.google.com/drive/folders/0B5eBgc8jrKtpUmhhSmtFLWdrTzQ}} 
 which obtains a  tagging accuracy at 93.0\%, a UAS score at 77.7\% and a LAS score at 69.5\% when evaluated on the Vietnamese dependency treebank VnDT of 10k sentences \cite{Nguyen2014NLDB}.

\section{Experimental methodology}

We compare the joint word segmentation and POS tagging strategy to the pipeline strategy on the benchmark Vietnamese treebank \cite{nguyen-EtAl:2009:LAW-III}  using well-known POS tagging models.

\subsection{Joint segmentation and POS tagging}

Following  \newcite{kruengkrai-EtAl:2009:ACLIJCNLP},  \newcite{Takahashi2016} and \newcite{ShaoHTN17}, we  formalize the joint  word segmentation and POS tagging problem for Vietnamese as a sequence labeling task to assigning a combined segmentation and POS tag to each syllable. For example, given a manually POS-annotated training corpus ``Cuộc/Nc điều\_tra/V dường\_như/X không/R tiến\_triển/V \textbf{.}/CH'' `The investigation seems to be making no progress', we transform this corpus into a syllable-based representation as follows: ``Cuộc/B-Nc điều/B-V tra/I-V dường/B-X như/I-X không/B-R tiến/B-V triển/I-V \textbf{.}/B-CH'', where segmentation tags B and I denote beginning  and inside of a word, respectively, while Nc, V, X, R and CH are POS tags. Then we train sequence labeling models on the syllable-based transformed corpus.

\subsection{Dataset}

The Vietnamese treebank \cite{nguyen-EtAl:2009:LAW-III} is the largest annotated corpus for Vietnamese,  providing a set of 27,870 manually POS-annotated sentences for training and development  (about 23 words per
sentence on average) and  a test set of 2120 manually POS-annotated sentences (about 31 words per sentence).\footnote{The data was officially used for the Vietnamese POS tagging shared task at the second VLSP 2013 workshop.} From the set of 27,870 sentences, we use the first 27k sentences for training and the last 870 sentences for development.

\subsection{Models} \label{ssec:models}

For both joint and pipeline strategies, we use the following models: 

\begin{itemize}
    \item {RDRPOSTagger} \cite{NguyenEACL2014NPP} is a transformation rule-based learning model which obtained the highest accuracy at the VLSP 2013 POS tagging shared task.\footnote{\url{http://rdrpostagger.sourceforge.net}} 
    
    \item {MarMoT} \cite{mueller-schmid-schutze:2013:EMNLP} is a generic CRF framework and a SOTA POS and morphological tagger.\footnote{\url{http://cistern.cis.lmu.de/marmot}}
    
    \item {BiLSTM-CRF} \cite{HuangXY15}  is a sequence labeling model which extends the BiLSTM model with a CRF layer.

    \item  {BiLSTM-CRF + CNN-char}, i.e. {BiLSTM-CNN-CRF}, is an extension of the {BiLSTM-CRF}, using CNN to derive character-based representations   \cite{ma-hovy:2016:P16-1}.%\footnote{\url{https://github.com/XuezheMax/LasagneNLP}}
    
    \item {BiLSTM-CRF + LSTM-char}   is another extension of the {BiLSTM-CRF}, using BiLSTM to derive the character-based representations \cite{lample-EtAl:2016:N16-1}.

\end{itemize}

Here, for the pipeline strategy, we train these models to predict POS tags with respect to  (w.r.t.) gold  word segmentation. In addition, we also retrain the fast and accurate Vietnamese word segmenter {RDRsegmenter} \cite{NguyenNVDJ2017} using the training set of 27k sentences.\footnote{RDRsegmenter obtains a segmentation speed at
60k words per second, computed  on a personal computer of Intel Core
i7 2.2 GHz. RDRsegmenter is available at: \url{https://github.com/datquocnguyen/RDRsegmenter}}

\subsection{Implementation details}

We use the original pure Java  implementations of {RDRPOSTagger} and {MarMoT} with default hyper-parameter settings in our experiments. Instead of using   implementations independently provided  by authors of  {BiLSTM-CRF}, {BiLSTM-CRF + CNN-char}\footnote{\url{https://github.com/XuezheMax/LasagneNLP}} and {BiLSTM-CRF + LSTM-char}, we use a reimplementation which  is optimized for performance of all these models from \newcite{reimers-gurevych:2017:EMNLP2017}.\footnote{\url{https://github.com/UKPLab/emnlp2017-bilstm-cnn-crf}}

For three {BiLSTM-CRF}-based models, we use default hyper-parameters provided by \newcite{reimers-gurevych:2017:EMNLP2017}  with the
following exceptions: we use a dropout rate at 0.5 \cite{ma-hovy:2016:P16-1} with the frequency threshold of 5 for unknown word and syllable types. We initialize word and syllable embeddings  with 100-dimensional pre-trained embeddings,\footnote{Pre-trained word and syllable embeddings are learned by training the Word2Vec Skip-gram model \cite{NIPS2013_5021} on a Vietnamese news corpus which is available  at: \url{http://mim.hus.vnu.edu.vn/phuonglh/corpus/baomoi.zip}} then learn them  together with other model parameters during training by using Nadam \cite{Dozat2015IncorporatingNM}. For training, we run for 100 epochs. We perform a grid search of hyper-parameters  to select the number of BiLSTM layers from $\{1, 2, 3\}$ and the number of LSTM units in each layer from $\{50, 100, 150, 200, 250, 300\}$. 
Early stopping is applied when no performance improvement on the development set is obtained after 5 contiguous epochs. 
For both pipeline and joint strategies, we find the highest performance on the development set is when using two stacked BiLSTM layers.  Table \ref{tab:optimal} presents the optimal number of LSTM units.

\begin{table}[!t]
    \centering
    \begin{tabular}{l|c|l }
    \hline
     \textbf{Model}  & \textbf{Pipeline} & \textbf{Joint}  \\
     \hline
     {BiLSTM-CRF} &  100 & 200  \\
      {\ \ \ \ \ + CNN-char} &  100 & 250  \\
       {\ \ \ \ \ + LSTM-char} &  150 &  250  \\
    \hline
    \end{tabular}
    \caption{Optimal number of LSTM units.}
    \label{tab:optimal}
\end{table}

Here the performance is evaluated by F1 score, based on the number of correctly segmented and tagged words \cite{zhang-clark:2008:ACLMain}. In the case of gold word segmentation, F1 score for POS tagging is in fact the tagging accuracy.

 \section{Main results}
 
 Table \ref{tab:pos} presents POS tagging accuracy and tagging speed of each model on the test set w.r.t. gold word segmentation, in which {MarMoT} is the most accurate model while {RDRPOSTagger} is the fastest one. In particular, {MarMoT} obtains 0.5\%+ higher accuracy than the three BiLSTM-based models.  This is not surprising as the training set of 27k sentences is relatively small compared to the training data available in other languages such as English or Chinese.
 
Table \ref{tab:wordseg} presents F1 scores  for word segmentation  and POS tagging in a real-world application scenario where the gold word-segmentation is not available.  Comparing the results in  Table \ref{tab:pos}  to  results for the pipeline strategy, we observe a drop of about 2\% for all models when using predicted  segmentation instead of gold segmentation. Also, Table \ref{tab:wordseg}  clearly shows that the pipeline strategy helps produce better results than the joint strategy.  
In addition,  pre-designed features in  both {RDRPOSTagger} and {MarMoT} are designed to capture word-level information rather than syllable-level information,  so it is also not surprising that for the joint strategy  {RDRPOSTagger} is significantly lower while {MarMoT}  is  lower  than the {BiLSTM-CRF} model with additional character-based representations.

\begin{table}[!t]
    \centering
    \begin{tabular}{l|c|l }
    \hline
     \textbf{Model}  & \textbf{Accuracy} & \textbf{Speed}  \\
     \hline
     {RDRPOSTagger} & 95.11  & \textbf{180k} \\
        {MarMoT} & \textbf{95.88}  & 25k \\
      {BiLSTM-CRF} &  95.06 & 3k  \\
      {\ \ \ \ \ + CNN-char} &  95.40 &  2.5k  \\
       {\ \ \ \ \ + LSTM-char} &  95.31 &  1.5k   \\
    \hline
    \end{tabular}
    \caption{POS tagging accuracies (in \%) on the test set w.r.t. gold word segmentation. ``Speed'' denotes the tagging speed, i.e. the number of words per second, computed on a personal computer of Intel Core i7 2.2 GHz (model loading time is not taken into account).}
    \label{tab:pos}
\end{table}

\begin{table}[!t]
    \centering
    \begin{tabular}{c|l|c|c}
    \hline
     \multicolumn{2}{c|}{\textbf{Model}} & \textbf{WSeg} & \textbf{PTag} \\
     \hline
    \multirow{5}{*}{\rotatebox[origin=c]{90}{Pipeline}}  
        & {RDRPOSTagger} & {97.75}  &  93.39 \\
        & {MarMoT} & {97.75}  &  \textbf{93.96} \\
        & {BiLSTM-CRF} &  {97.75} & 93.25 \\
        & {\ \ \ \ \ + CNN-char} &  {97.75} & 93.55 \\
        & {\ \ \ \ \ + LSTM-char} &  {97.75} & 93.46 \\
    \hline
    \multirow{5}{*}{\rotatebox[origin=c]{90}{Joint}}  
        & {RDRPOSTagger} & 93.73  &  87.53 \\
        & {MarMoT} & 96.50  & 92.78 \\
        & {BiLSTM-CRF} & 96.15  & 92.43 \\
        & {\ \ \ \ \ + CNN-char} & 96.66   & 92.79 \\
        & {\ \ \ \ \ + LSTM-char} &  96.76 &  92.95 \\
    \hline
    \end{tabular}
    \caption{ F1 scores (in \%)  for word segmentation (\textbf{WSeg}) and POS tagging (\textbf{PTag}) from unsegmented  text. The  pipeline strategy uses {RDRsegmenter} for word segmentation. In preliminary experiments, where we also train the five models above to predict a segmentation tag B or I for  each syllable, we then find that {RDRsegmenter} obtains better word segmentation score than those five models.}
    \label{tab:wordseg}
\end{table}

Tables \ref{tab:pos} and \ref{tab:wordseg} suggest that for a practical application to Vietnamese where  performance accuracy is preferred, we should consider using the pipeline strategy with a traditional SOTA feature-based tagger such as {MarMoT}. If   speed is  preferred such as in big data, {RDRPOSTagger} would be a superior alternative. With the current state of  training data available in Vietnamese, future research should focus on incorporating Vietnamese linguistic features into the traditional feature-based sequence taggers.

\section{Conclusion}

We have presented empirical comparisons between two strategies for Vietnamese POS tagging from unsegmented text  and between SOTA feature- and neural network-based models. Experimental
results on  the benchmark Vietnamese treebank \cite{nguyen-EtAl:2009:LAW-III} show that the pipeline strategy produces  higher scores of POS tagging  from unsegmented text than the joint strategy. In addition, we also show that a traditional feature-based model (i.e. MarMoT) obtains better POS tagging accuracy than neural network-based models. 
We provide a pre-trained MarMoT model for Vietnamese POS tagging at \url{https://github.com/datquocnguyen/VnMarMoT}.

\section*{Acknowledgments}
This research was partially supported by the Australian Government through the Australian Research Council's Discovery Projects funding scheme (project DP160102156). This research was also partially supported by NICTA, funded by the Australian Government through the Department of Communications and the Australian Research Council through the ICT Centre of Excellence Program. 

\bibliographystyle{acl_natbib}
\bibliography{refs}

\end{document}